\documentclass[11pt,a4paper]{article}
\usepackage[hyperref]{acl2019}
\usepackage{times}
\usepackage{latexsym}

\usepackage{url}

\usepackage{amsmath}
\usepackage{bm}
\usepackage{color}
\usepackage{multirow}
\usepackage{subfig}
\usepackage{arydshln}
\usepackage{url}
\usepackage{amsfonts}
\usepackage{graphicx} 
\usepackage{graphics}

\usepackage{caption}
\usepackage{balance}

\aclfinalcopy 
\def\aclpaperid{2112} 

\definecolor{electricviolet}{rgb}{0.56, 0.0, 1.0}

\title{Assessing the Ability of Self-Attention Networks to Learn Word Order}

\author{Baosong Yang$^\dagger$~~~~~Longyue Wang$^\ddagger$~~~~~Derek F. Wong$^{\dagger}$~~~~~Lidia S. Chao$^\dagger$~~~~~Zhaopeng Tu$^\ddagger$\thanks{~~~Zhaopeng Tu is the corresponding author of the paper. This work was conducted when Baosong Yang was interning at Tencent AI Lab.}\\
  $^\dagger$NLP$^2$CT Lab, Department of Computer and Information Science,
  University of Macau\\
  {\tt nlp2ct.baosong@gmail.com, \{derekfw,lidiasc\}@umac.mo} \\
  $^\ddagger$Tencent AI Lab\\
  {\tt \{vinnylywang,zptu\}@tencent.com}}

\begin{document}
\maketitle
\begin{abstract}

Self-attention networks (SAN) 
have attracted a lot of interests 
due to their high parallelization and strong performance on a variety of NLP tasks, e.g. machine translation. 
Due to the lack of recurrence structure such as recurrent neural networks (RNN), SAN is ascribed to be weak at learning positional information of words for sequence modeling. 
However, neither this speculation has been empirically confirmed, nor explanations for their strong performances on machine translation tasks when ``lacking positional information'' have been explored. 
To this end, we propose a novel {\em word reordering detection} task to 
quantify how well the word order information learned by SAN and RNN. Specifically, we randomly move one word to another position, and examine whether a trained model can detect both the original and inserted positions.
Experimental results reveal that: 1) SAN trained on {\em word reordering detection} indeed has difficulty learning the positional information even with the position embedding; and 2) SAN trained on {\em machine translation} learns better positional information than its RNN counterpart, in which position embedding plays a critical role. Although recurrence structure make the model more universally-effective on learning word order, learning objectives matter more in the downstream tasks such as machine translation.
\end{abstract}

\section{Introduction}
\label{sec:intro}
Self-attention networks~\citep[{SAN},][]{parikh2016decomposable,lin2017structured} have shown promising empirical results in a variety of natural language processing (NLP) tasks, such as machine translation~\cite{Vaswani:2017:NIPS}, 
semantic role labelling~\cite{strubell2018linguistically}, and language representations~\cite{devlin2018bert}.
The popularity of SAN lies in its high parallelization in computation, and flexibility in modeling dependencies regardless of distance by explicitly attending to all the signals.
Position embedding~\cite{pmlr-v70-gehring17a} is generally deployed to capture sequential information for SAN~\cite{Vaswani:2017:NIPS,Shaw:2018:NAACL}.

Recent studies claimed that SAN with position embedding is still weak at learning word order information, due to the lack of recurrence structure that is essential for sequence modeling~\cite{Shen:2018:AAAI,Chen:2018:ACL,Hao:2019:NAACL}.
However, such claims are mainly based on a theoretical argument, which have not been empirically validated. In addition, this can not explain well why SAN-based models outperform their RNN counterpart in machine translation -- a benchmark sequence modeling task~\cite{Vaswani:2017:NIPS}.

\begin{figure*}[t]
\begin{center}
\subfloat[Position Detector]{\includegraphics[width=0.55\textwidth]{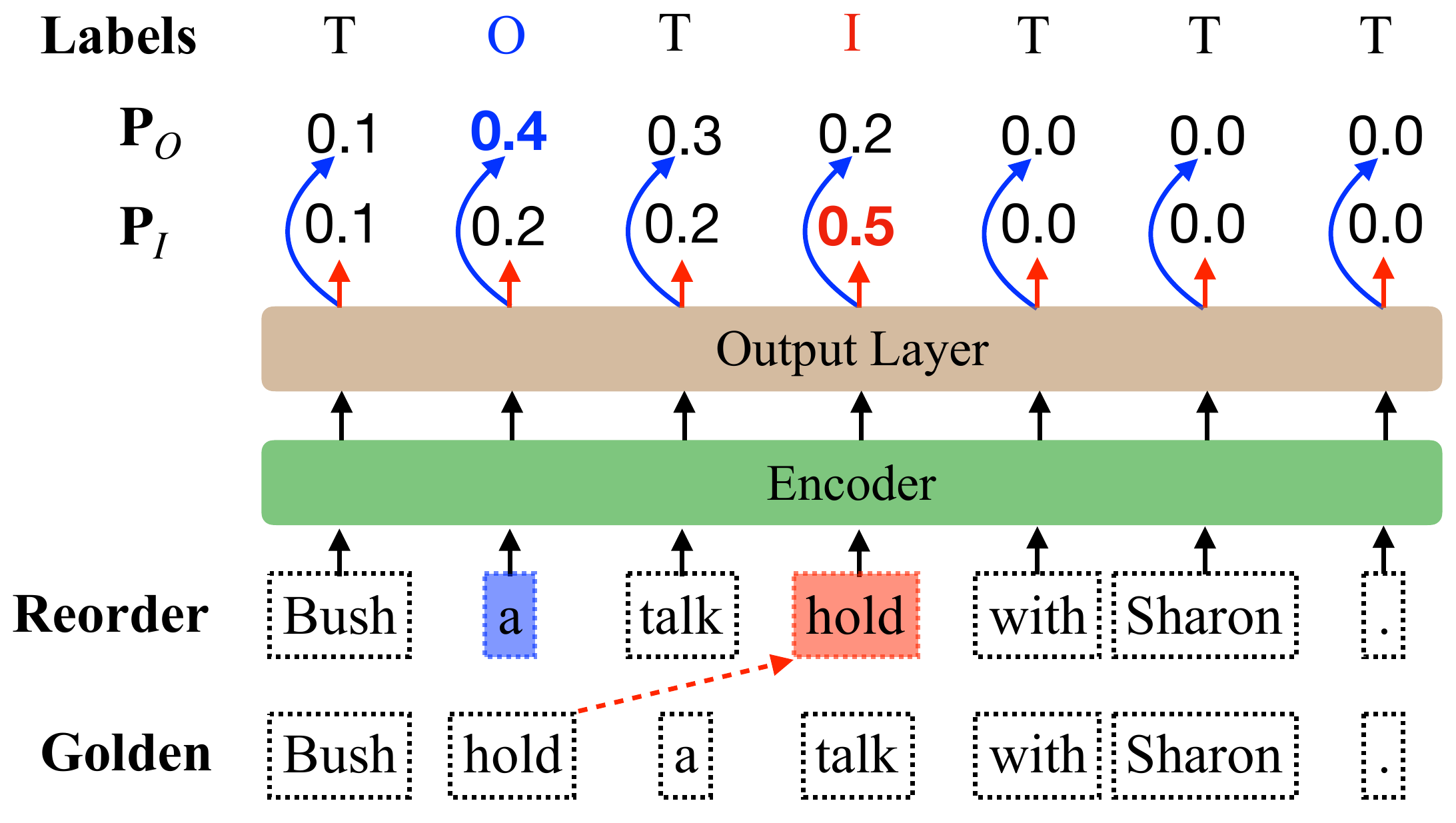}} \hspace{0.1\textwidth}
\subfloat[Output Layer]{\includegraphics[width=0.20\textwidth]{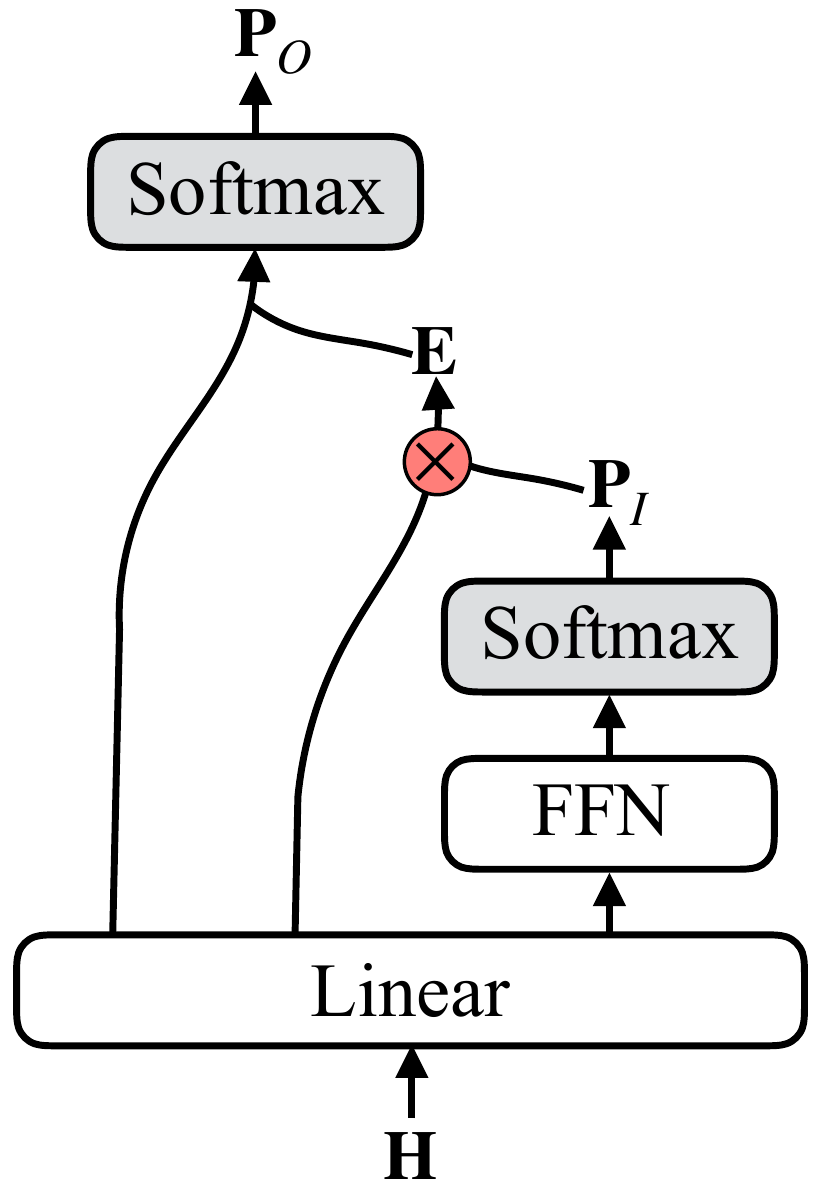}}
\caption{Illustration of (a) the position detector, where (b) the output layer is build upon a randomly initialized or pre-trained encoder. In this example, the word ``hold'' is moved to another place. The goal of this task is to predict the inserted position ``I'' and the original position ``O'' of ``hold''.}
\label{fig:cls}
\end{center}
\end{figure*}

Our goal in this work is to empirically assess the ability of SAN to learn word order. 
We focus on asking the following research questions:
\begin{itemize}
    \item[{\bf Q1}:] Is recurrence structure obligate for learning word order, and does the conclusion hold in different scenarios (e.g., translation)?
    \item[{\bf Q2}:] Is the model architecture the critical factor for learning word order in the downstream tasks such as machine translation?
    \item[{\bf Q3}:] Is position embedding powerful enough to capture word order information for SAN?
\end{itemize}

We approach these questions with a novel probing task -- {\em word reordering detection} (WRD), which aims to detect the positions of randomly reordered words in the input sentence. We compare SAN with RNN, as well as directional SAN ~\citep[{DiSAN},][]{Shen:2018:AAAI} that augments SAN with recurrence modeling. In this study, we focus on the encoders implemented with different architectures, so as to investigate their abilities to learn word order information of the input sequence. The encoders are trained on objectives like detection accuracy and machine translation, to study the influences of learning objectives.

Our experimental results reveal that: (Q1) SAN indeed underperforms the architectures with recurrence modeling (i.e. RNN and DiSAN) on the WRD task, while this conclusion does not hold in machine translation: SAN trained with the translation objective outperforms both RNN and DiSAN on detection accuracy;
(Q2) Learning objectives matter more than model architectures in downstream tasks such as machine translation; and (Q3) Position encoding is good enough for SAN in machine translation, while DiSAN is a more universally-effective mechanism to learn word order information for SAN.


\paragraph{Contributions}
The key contributions are:
\begin{itemize}
    \item We design a novel probing task along with the corresponding benchmark model, which can assess the abilities of different architectures to learn word order information.\footnote{The data and codes are released at:  \url{https://github.com/baosongyang/WRD}.} 
    \item Our study dispels the doubt on the inability of SAN to learn word order information in machine translation, indicating that the learning objective can greatly influence the suitability of an architecture 
    for downstream tasks.
\end{itemize}

\section{Word Reordering Detection Task}
\label{sec:order}
In order to investigate the ability of self-attention networks to extract  word order information, in this section, we design an artificial task to evaluate the abilities of the examined models to detect the erroneous word orders in a given sequence.

\paragraph{Task Description}

Given a sentence $X=\{x_1,...,x_i,...,x_N\}$, we randomly pop a word $x_i$ and insert it into another position $j$ ($1\leq i,j \leq N$ and $i\neq j$). 
The objective of this task is to detect both the position the word is popped out (labeled as ``O''), as well as the position the word is inserted (labeled as ``I''). 
As seen the example in Figure~\ref{fig:cls} (a), 
the word ``hold'' is moved from the $2^{nd}$ slot to the $4^{th}$ slot. Accordingly, the $2^{nd}$ and $4^{th}$ slots are labelled as ``O'' and ``I'', respectively. 
{To exactly detect word reordering, the examined models have to learn to  recognize both the normal and abnormal word order in a sentence.}

\paragraph{Position Detector}
Figure~\ref{fig:cls} (a) depicts the architecture of the position detector.
Let the sequential representations ${\bf H} = \{{\bf h}_1,..., {\bf h}_N\}$ be the output of each encoder noted in Section~\ref{sec:bg}, which are fed to the output layer (Figure~\ref{fig:cls} (b)). Since only one pair of ``I'' and ``O'' labels should be generated in the output sequence, we cast the task as a pointer detection problem~\cite{vinyals2015pointer}. To this end, we turn to an output layer that commonly used in the reading comprehension task~\cite{wang2016machine,du2017identifying}, which aims to identify the start and end positions of the answer in the given text.\footnote{Contrary to reading comprehension in which the start and end positions are ordered, ``I'' and ``O'' do not have to be ordered in our tasks, that is, the popped word can be inserted to either left or right position.}
The output layer consists of two sub-layers, which progressively predicts the probabilities of each position being labelled as ``I'' and ``O''.
The probability distribution of the sequence being labelled as ``I'' is calculated as:
\begin{align}
{\bf P}_I &= {\tt SoftMax}({\bf U}_I^\top{\tt tanh}({\bf W}_I {\bf H}))      &&    \in \mathbb{R}^N
\end{align}
where ${\bf W}_I\in \mathbb{R}^{d \times d} $ and ${\bf U}_I\in \mathbb{R}^d$ are trainable parameters, and $d$ is the dimensionality of $\bf H$.

The second layer aims to locate the original position ``O'', which conditions on the predicted popped word at the position ``I''.\footnote{We tried to predict the position of ``O'' without feeding the approximate embedding, i.e. predicting ``I'' and ``O'' individually. It slightly underperforms the current model.}
To make the learning process differentiable, we follow \newcite{xu2017neural} to use the weighted sum of hidden states as the approximate embedding $\bf E$ of the popped word. The embedding subsequently serves as a query to attend to the sequence $\bf H$ to find which position is most similar to the original position of popped word. The probability distribution of the sequence being labelled as ``O'' is calculated as:
\begin{align}
\bf E &= {\bf P}_I ({\bf W}_Q {\bf H})     &&    \in \mathbb{R}^d\\
{\bf P}_O &= \textsc{Att}({\bf E}, {\bf W}_K {\bf H})    &&    \in \mathbb{R}^N
\end{align}
where $\{{\bf W}_Q, {\bf W}_K\} \in \mathbb{R}^{d \times d}$ are trainable parameters that transform $\bf H$ to query and key spaces respectively.  $\textsc{Att}(\cdot)$ denotes the dot-product attention~\cite{Luong:2015:EMNLP,Vaswani:2017:NIPS}. 

\paragraph{Training and Predicting} 
In training process, the objective is to minimize the cross entropy of the true inserted and original positions, which is the sum of the negative log probabilities of the groundtruth indices by the predicted distributions:
\begin{eqnarray}
L = {\bf Q}^\top_I \log {\bf P}_I + {\bf Q}^\top_O \log {\bf P}_O
\end{eqnarray}
where $\{{\bf Q}_I, {\bf Q}_O\} \in \mathbb{R}^{N}$ is an one-hot vector to indicate the groundtruth indices for the inserted and original positions.
During prediction, we choose the positions with highest probabilities from the distributions ${\bf P}_I$ and ${\bf P}_O$ as ``I'' and ``O'', respectively. Considering the instance in Figure~\ref{fig:cls} (a), the $4^{th}$ position is labelled as inserted position ``I'', and the $2^{nd}$ position as the original position ``O''.

\section{Experimental Setup}
\label{sec:bg}
In this study, we strove to empirically test whether SAN indeed  weak at learning positional information and come up with the reason about the strong performance of SAN on machine translation. In response to the three research questions in Section~\ref{sec:intro}, we give following experimental settings:
\begin{itemize}
    \item Q1: We compare SAN with two recurrence architectures -- RNN and DiSAN on the WRD task, thus to quantify their abilities on learning word order (Section~\ref{sec:enc}).
    \item Q2: To compare the effects of learning objectives and model architectures, we train each encoder under two scenarios, i.e. trained on objectives like WRD accuracy and on machine translation (Section~\ref{sec:train}).
    \item Q3: The strength of position encoding is appraised by ablating position encoding and recurrence modeling for SAN. 
\end{itemize}

\subsection{Encoder Setting}
\label{sec:enc}
\begin{figure}[h]
\begin{center}
\subfloat[RNN]{\includegraphics[width=0.29\columnwidth]{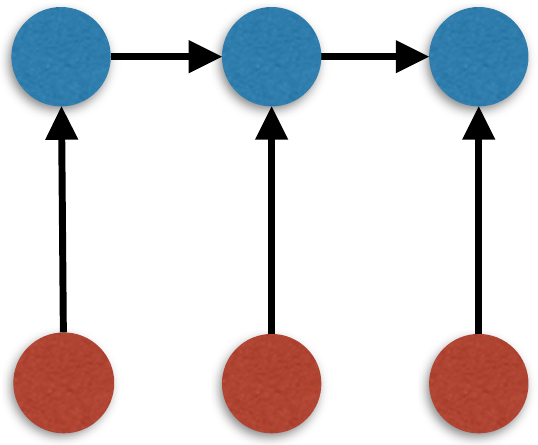}}\hspace{0.05\columnwidth}
\subfloat[SAN]{\includegraphics[width=0.29\columnwidth]{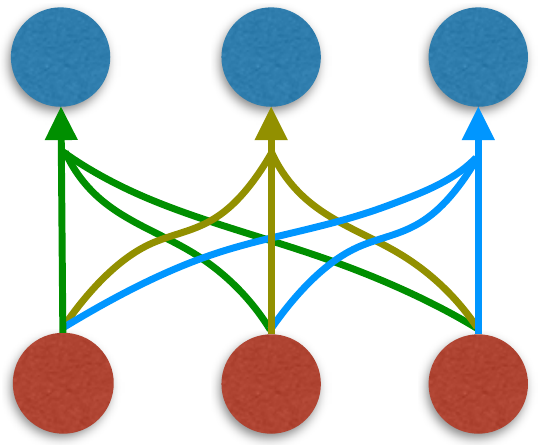}}\hspace{0.05\columnwidth}
\subfloat[DiSAN]{\includegraphics[width=0.29\columnwidth]{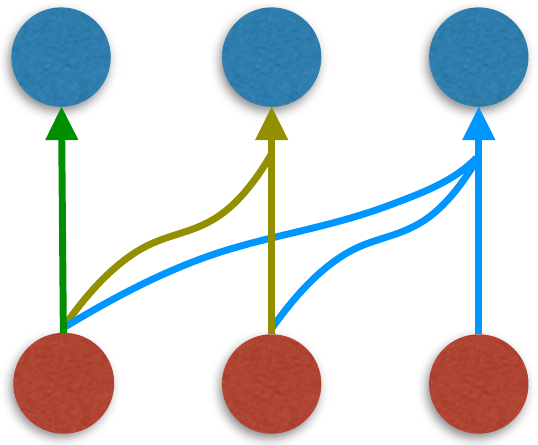}}
\caption{Illustration of (a) RNN; (b) SAN; and (c) DiSAN. Colored arrows denote parallel operations.}
\label{fig:models}
\end{center}
\end{figure}

\noindent RNN and SAN are commonly used to produce sentence representations on NLP tasks~\cite{cho2014learning,lin2017structured,Chen:2018:ACL}.  
As shown in Figure~\ref{fig:models}, we investigate three architectures in this study. 
Mathematically, let ${\bf X}=\{{\bf x}_1, \dots, {\bf x}_N\} \in \mathbb{R}^{d \times N}$ be the embedding matrix of the input sentence, and ${\bf H}=\{{\bf h}_1, \dots, {\bf h}_N\} \in \mathbb{R}^{d \times N}$ be the output sequence of representations.

\begin{itemize}
  \item {\bf RNN} sequentially produces each state:
    \begin{eqnarray}
        {\bf h}_n = f({\bf h}_{n-1}, {\bf x}_n),
    \end{eqnarray}
    where $f(\cdot)$ is GRU~\cite{cho2014learning} in this study. 
    RNN is particularly hard to parallelize due to their inherent dependence on the previous state ${\bf h}_{n-1}$.
  \item {\bf SAN}~\cite{lin2017structured} produces each hidden state in a parallel fashion:
    \begin{eqnarray}
        {\bf h}_n = \textsc{Att}({\bf q}_n, {\bf K}) {\bf V},
    \end{eqnarray}
    where the query ${\bf q}_n \in \mathbb{R}^d$ and the keys and values $({\bf K}, {\bf V}) \in \mathbb{R}^{d \times N}$ are transformed from ${\bf X}$. 
    To imitate the
order of the sequence, \newcite{Vaswani:2017:NIPS} deployed position encodings~\cite{pmlr-v70-gehring17a} into SAN.  
  \item {\bf DiSAN}~\cite{Shen:2018:AAAI} augments SAN with the ability to encode word order: 
    \begin{eqnarray}
        {\bf h}_n = \textsc{Att}({\bf q}_n, {\bf K}_{\leq n}) {\bf V}_{\leq n},
    \end{eqnarray}
    where $({\bf K}_{\leq n}, {\bf V}_{\leq n})$ indicate leftward elements, e.g., ${\bf K}_{\leq n} = \{{\bf k}_1, \dots, {\bf k}_n\}$.
\end{itemize}

To enable a fair comparison of architectures, we only vary the sub-layer of sequence modeling (e.g. the SAN sub-layer) in the Transformer encoder~\cite{Vaswani:2017:NIPS}, and keep the other components the same for all architectures. We use bi-directional setting for RNN and DiSAN, and apply position embedding for SAN and DiSAN. 
We follow \newcite{Vaswani:2017:NIPS} to set the configurations of the encoders, which consists of 6 stacked layers with the layer size being 512.

\subsection{Learning Objectives}
\label{sec:train}

In this study, we employ two strategies to train the encoders, which differ at the learning objectives and data used to train the associated parameters. Note that in both strategies, the output layer in Figure~\ref{fig:models} is fine-trained on the WRD data with the word reordering detection objective.


\paragraph{WRD Encoders} We first directly train the encoders on the WRD data, to evaluate the abilities of model architectures. The WRD encoders are randomly initialized and co-trained with the output layer. Accordingly, the detection accuracy can be treated as the learning objective of this group of encoders. Meanwhile, we can investigate the reliability of the proposed WRD task by checking whether the performances of different architectures (i.e. RNN, SAN, and DiSAN) are consistent with previous findings on other benchmark NLP tasks~\cite{Shen:2018:AAAI,Tang:2018:EMNLP,tran2018importance,devlin2018bert}.

\paragraph{NMT Encoders} To quantify how well different architectures learn word order information with the learning objective of machine translation, we first train the NMT models (both encoder and decoder) on bilingual corpus using the same configuration reported by \newcite{Vaswani:2017:NIPS}. Then, we {\em fix the parameters of the encoder}, and only train the parameter associated with the output layer on the WRD data. In this way, we can probe the representations learned by NMT models, on their abilities to learn word order of input sentences.

To cope with WRD task, all the models were trained for 600K steps, each of which is allocated a batch of 500 sentences. The training set is shuffled after each epoch. We use Adam~\cite{kingma2015adam} with $\beta_1 = 0.9$, $\beta_2 = 0.98$ and $\epsilon = 10^{-9}$. The learning rate linearly warms up over the first 4,000 steps, and decreases thereafter proportionally to the inverse square root of the step number. We use a dropout rate of 0.1 on all layers.


\subsection{Data}
\label{sec:data}

\paragraph{Machine Translation}
We pre-train NMT models on the benchmark WMT14 English$\Rightarrow$German (En$\Rightarrow$De) data, which consists of 4.5M sentence pairs. The validation and test sets are newstest2013 and newstest2014, respectively.
To demonstrate the universality of the findings in this study, we also conduct experiments on WAT17 English$\Rightarrow$Japanese (En$\Rightarrow$Ja) data. Specifically, we follow~\newcite{morishita2017ntt} to use the first two sections of WAT17 dataset as the training data, which approximately consists of 2.0M sentence pairs. We use newsdev2017 as the validation set and newstest2017 as the test set.

\paragraph{Word Reordering Detection}
We conduct this task on the English sentences, which are extracted from the source side of WMT14 En$\Rightarrow$De data with maximum length to 80.
For each sentence in different sets (i.e. training, validation, and test sets), we construct an instance by randomly moving a word to another position.
Finally we construct 7M, 10K and 10K samples for training, validating and testing, respectively. Note that a sentence can be sampled multiple times, thus each dataset in the WRD data contains more instances than that in the machine translation data.


All the English and German data are tokenized using the scripts in Moses. 
The Japanese sentences are segmented by the word segmentation toolkit {KeTea}~\cite{neubig2011pointwise}.
To reduce the vocabulary size, all the sentences are processed by byte-pair encoding (BPE)  \cite{sennrich2016neural} with 32K merge operations for all the data.

\section{Experimental Results}
\label{sec:exp}
We return to the central questions originally posed, that is, whether SAN is indeed weak at learning positional information. Using the above experimental design, 
 we give the following answers: %
\begin{itemize}
    \item[{\bf A1}:] SAN-based encoder trained on the WRD data is indeed harder to learn positional information than the recurrence architectures (Section~\ref{sec:exp-individual}), while there is no evidence that SAN-based NMT encoders learns less word order information (Section~\ref{sec:mt-order-res});
    \item[{\bf A2}:] The learning objective plays a more crucial role on learning word order than the architecture in downstream tasks (Section~\ref{sec:ana});
    \item[{\bf A3}:] While the position encoding is powerful enough to capture word order information in machine translation, DiSAN is a more universally-effective mechanism (Table~\ref{tab:mt}).
\end{itemize}

\subsection{Results on WRD Encoders}
\label{sec:exp-individual}

\begin{table}[t]
  \centering
  \begin{tabular}{l|ccc}
       {\bf Models}  & {\bf Insert} & {\bf Original} & {\bf Both}\\  \hline \hline
       RNN   & 78.4 & \bf 73.4 & \bf 68.2 \\
       \hdashline
       SAN   & 73.2 & 66.0 & 60.1   \\
       DiSAN & {\bf 79.6} & 70.1 & 68.0 \\ 
  \end{tabular}
   \caption{Accuracy on the WRD task. ``Insert'' and ``Original'' denotes the accuracies of detecting the inserted and original positions respectively, and ``Both'' denotes detecting both positions.}  
 \label{tab:order}
\end{table}

We first check the performance of each WRD encoder on the proposed WRD task 
from two aspects: 1) WRD accuracy; and
 2) learning ability.
\begin{figure}[t]
\begin{center}
\includegraphics[width=0.43\textwidth]{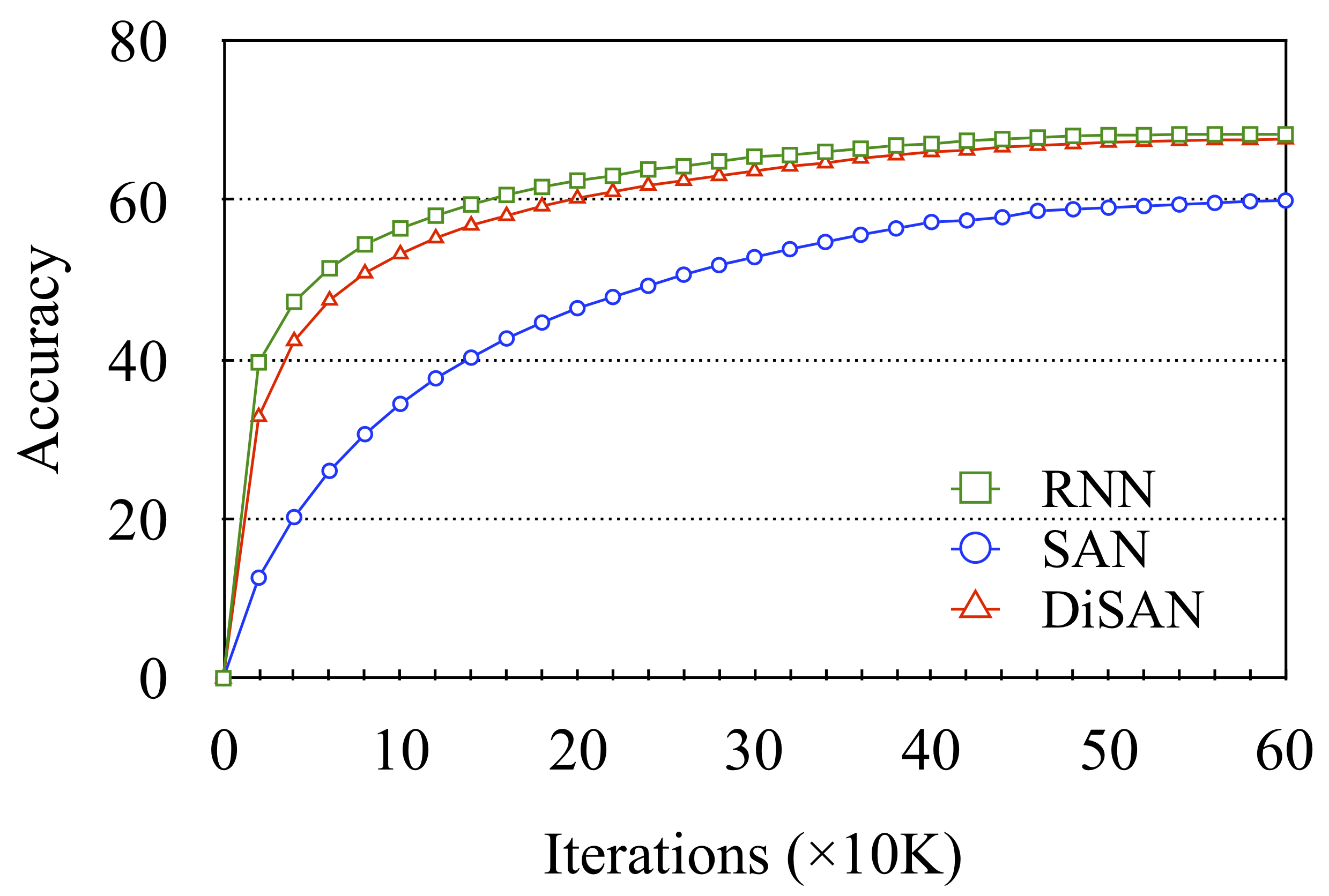}
\caption{Learning curve of WRD encoders on WRD task. Y-axis denotes the accuracy on the validation set. Obviously, SAN has slower convergence.}
\label{fig:learning}
\end{center}
\end{figure}

\begin{table*}[t]
\begin{center}
\begin{tabular}{l||cc||cc|c}
       \multirow{2}{*}{\bf Model}  & \multicolumn{2}{c||}{\bf Translation}  &  \multicolumn{3}{c}{\bf Detection}\\
    \cline{2-6}
        & En$\Rightarrow$De &   En$\Rightarrow$Ja    & En$\Rightarrow$De Enc. &   En$\Rightarrow$Ja  Enc.  &   WRD Enc.\\
    \hline \hline
       RNN	    &  26.8 & 42.9  & 33.9  &  29.0  &  \bf 68.2\\ 
    \hdashline
       SAN      &  27.3 & 43.6 & \bf 41.6  &\bf  32.8   &   60.1\\
       ~~- Pos\_Emb &  11.5 & -- & ~~0.3 & -- & ~~0.3   \\
       DiSAN    & \bf 27.6 &  \bf 43.7 & 39.7  &  31.2  &   68.0\\ 
       ~~- Pos\_Emb  &   27.0 & 43.1  & 40.1 & 31.0 & 62.8 \\ 
  \end{tabular}
  \caption{Performances of NMT encoders pre-trained on WMT14 En$\Rightarrow$De and WAT17 En$\Rightarrow$Ja data. ``Translation'' denotes translation quality measured in BLEU scores, while ``Detection'' denotes the accuracies on WRD task. ``En$\Rightarrow$De Enc.'' denotes NMT encoder trained with translation objective on the En$\Rightarrow$De data. We also list the detection accuracies of WRD encoders (``WRD Enc.'') for comparison. ``- Pos\_Emb'' indicates removing positional embeddings from SAN- or DiSAN-based encoder. Surprisingly, {\em SAN-based NMT encoder achieves the best accuracy on the WRD task}, which contrasts with the performances of WRD encoders (the last column).} 
  \label{tab:mt}
  \end{center}
\end{table*}

\paragraph{WRD Accuracy}
The detection results are concluded in Table~\ref{tab:order}. As seen, both RNN and DiSAN significantly outperform SAN on our task, 
indicating that the recurrence structure (RNN) exactly performs better than parallelization (SAN) on capturing word order information in a sentence. Nevertheless, the drawback can be alleviated by applying directional attention functions. The comparable result between DiSAN and RNN confirms the hypothesis by \newcite{Shen:2018:AAAI} and~\newcite{devlin2018bert} that directional SAN exactly improves the ability of SAN to learn word order. The consistency between prior studies and our results  verified the reliability of the proposed WRD task.

\paragraph{Learning Curve}
We visualize the learning curve of the training. As shown in Figure~\ref{fig:learning}, SAN has much slower convergence than others, showing that SAN has a harder time learning word order information than RNN and DiSAN. This is consistent with our intuition that the parallel structure is more difficult to learn word order information than those models with a sequential process. Considering DiSAN, although it has slightly slower learning speed at the early stage of the training, it is able to achieve comparable accuracy to RNN at the mid and late phases of the training.

\subsection{Results on Pre-Trained NMT Encoders}
\label{sec:mt-order-res}

\begin{figure*}[t]
\begin{center}
\subfloat[WRD  Encoder]{\includegraphics[width=0.3\textwidth]{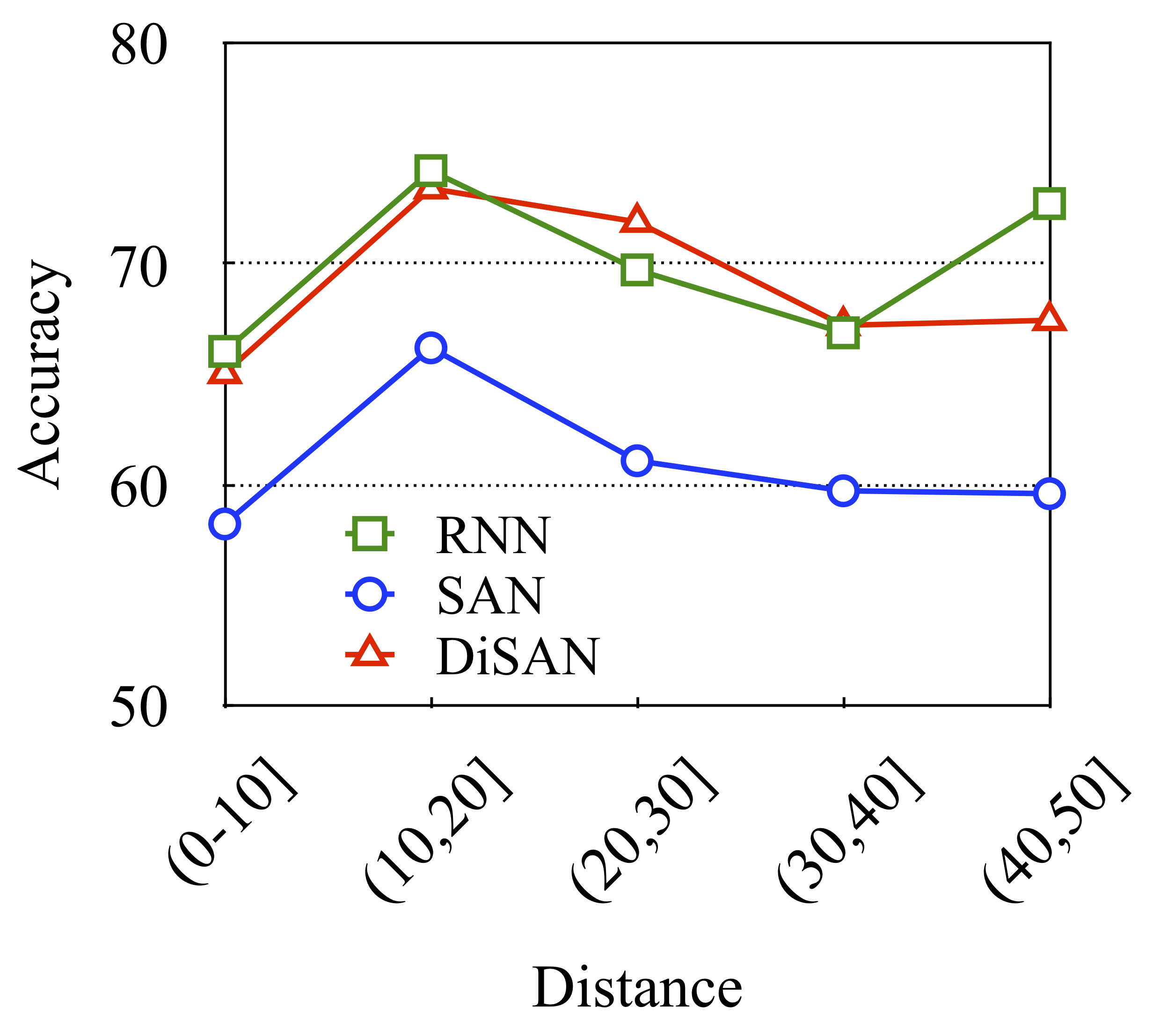}} \hspace{0.04\textwidth}
\subfloat[En$\Rightarrow$De NMT Encoder]{\includegraphics[width=0.3\textwidth]{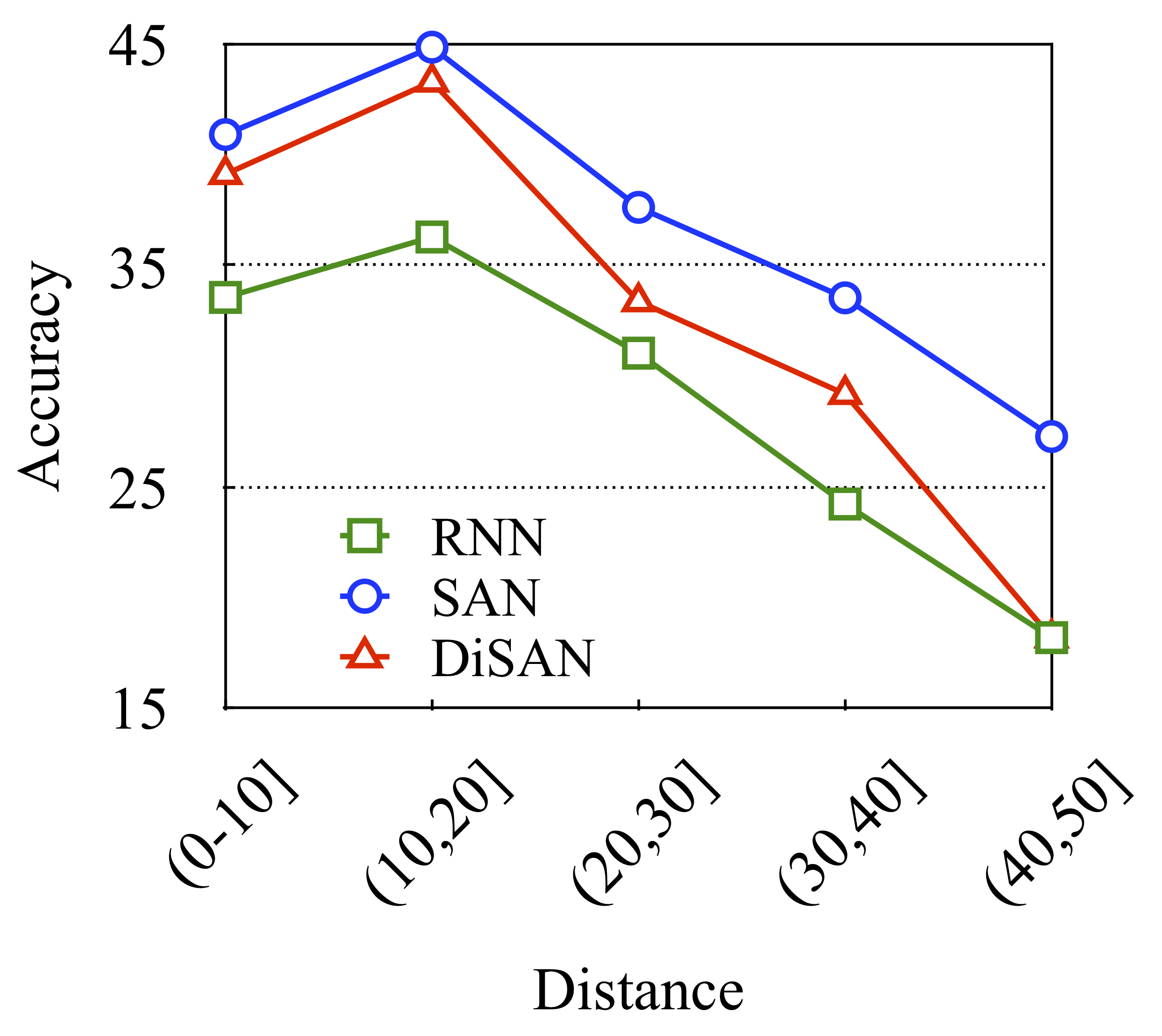}} \hspace{0.04\textwidth}
\subfloat[En$\Rightarrow$Ja NMT Encoder]{\includegraphics[width=0.3\textwidth]{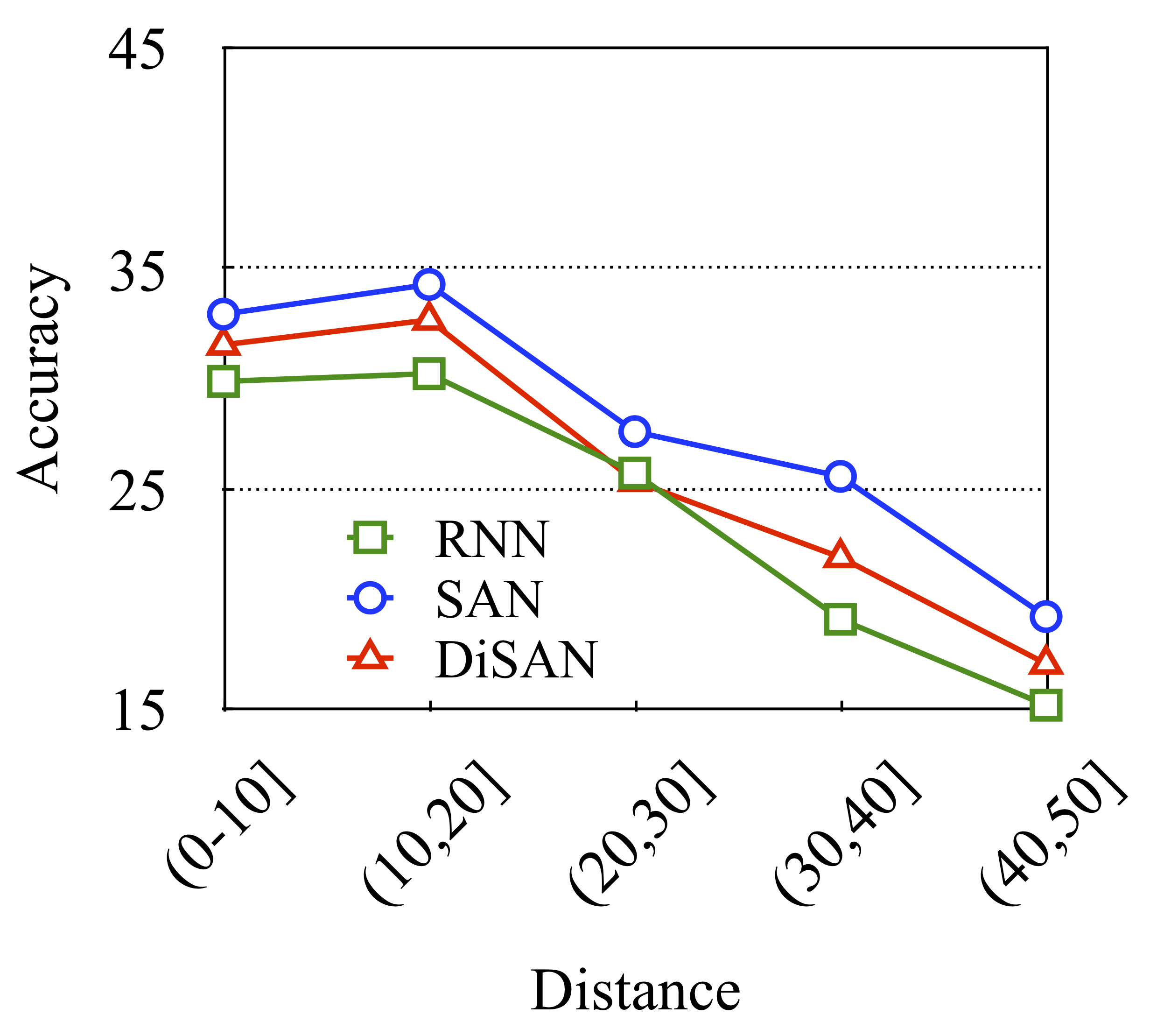}}
\caption{Accuracy of pre-trained NMT encoders according to various distances between the positions of ``O" and ``I'' (X-axis). As seen, the performance of each WRD encoder (a) is stable across various distances, while the pre-trained  (b) En$\Rightarrow$De and (c) En$\Rightarrow$Ja encoders consistently get lower accuracy with the increasing of distance.}
\label{fig:detect-dis}
\end{center}
\end{figure*}

\begin{figure*}[t]
\begin{center}
\subfloat[En$\Rightarrow$De NMT encoder]{\includegraphics[width=0.32\textwidth]{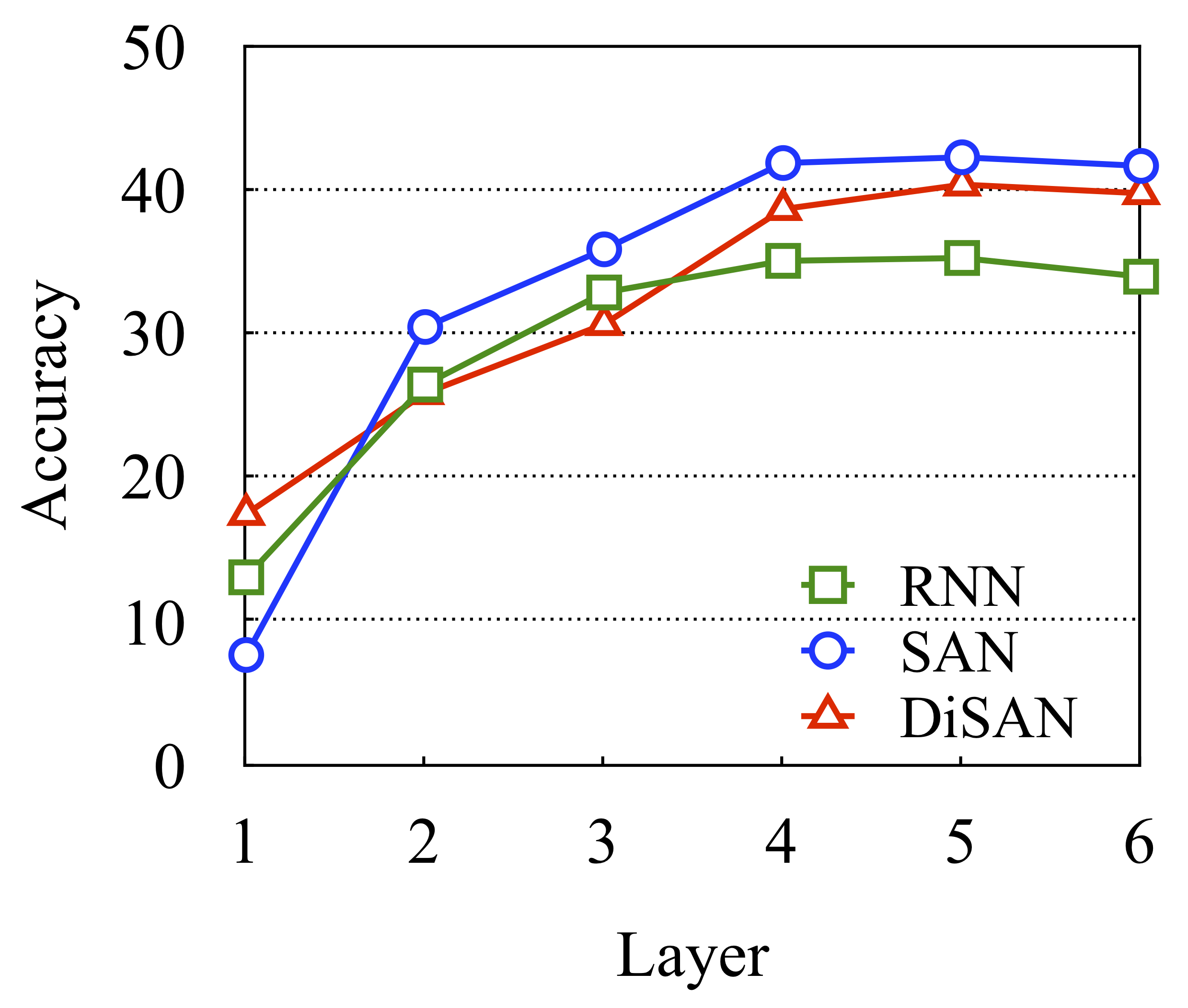}} \hspace{0.1\textwidth}
\subfloat[En$\Rightarrow$Ja NMT encoder]{\includegraphics[width=0.32\textwidth]{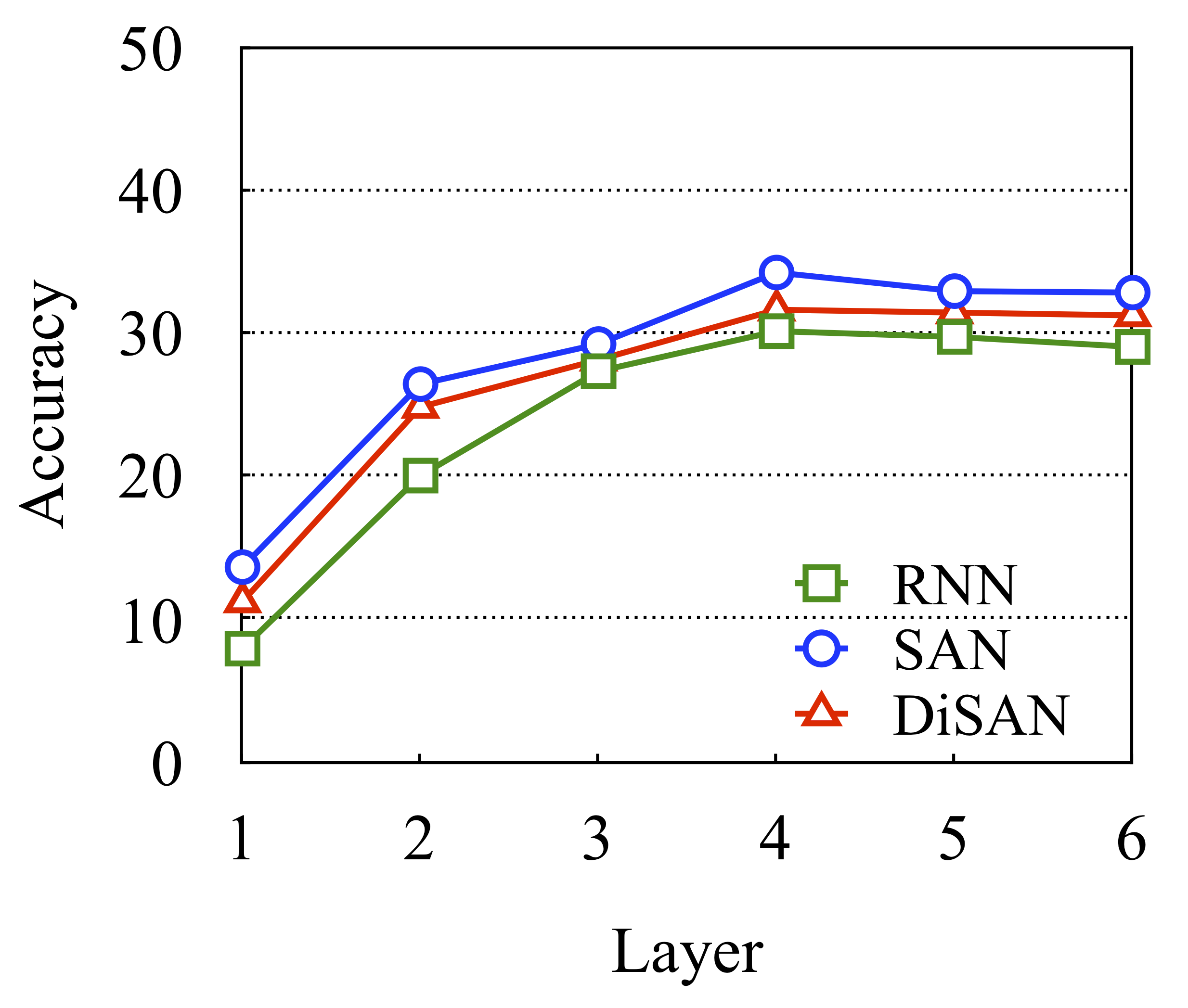}}
\caption{Performance of each layer from (a) pre-trained En$\Rightarrow$De encoder and (b) pre-trained En$\Rightarrow$Ja encoder  on WRD task. The evaluation are conducted on the test set. Clearly, the accuracy of SAN gradually increased with the stacking of layers and consistently outperform that of other models across layers.}
\label{fig:MT-layer}
\end{center}
\end{figure*}

We investigate whether the SAN indeed lacks the ability to learn word order information under machine translation context. The results are concluded in Table~\ref{tab:mt}. 
We first report the effectiveness of the compared models on translation tasks. 
For En-De translation, SAN outperforms RNN, which is consistent with the results reported in \cite{Chen:2018:ACL}. The tendency is universal on En-Ja which is a distant language pair \cite{bosch2001evidence,isozaki2010automatic}. 
Moreover, 
DiSAN incrementally improves the translation quality, demonstrating that model directional information benefits to the translation quality.
The consistent translation performances make the following evaluation on WRD accuracy convincing.

Concerning the performances of NMT encoders on the WRD task:
\paragraph{SAN-based NMT Encoder Performs Better}
It is surprising to see that SAN yields even higher accuracy on WRD task than other pre-trained NMT encoders, despite its lower translation qualities comparing with DiSAN. The results not only dispel the doubt on the inablity of SAN-based encoder to learn word order in machine translation, but also demonstrate that SAN learns to retain more features with respect to word order during the training of machine translation.

\paragraph{Learning Objectives Matter More}
In addition, both the NMT encoders underperform the WRD encoders on detection task across models and language pairs.\footnote{The En$\Rightarrow$Ja pre-trained encoders yield lower accuracy on WRD task than that of En$\Rightarrow$De pre-trained encoders. We attribute this to the difference between the source sentences in pre-training corpus (En-Ja) and that of WRD data (from En-De dataset). Despite of this, the tendency of results are consistent across language pairs.}  
The only difference between the two kinds of encoders is the learning objective. This raises a hypothesis that the learning objective sometimes severs as a more critical factor than the model architecture on modeling word order.

\paragraph{Position Encoding~VS.~Recurrence Modeling}
In order to assess the importance of position encoding, we redo the experiments by removing the position encoding from SAN and DiSAN (``- Pos\_Emb''). Clearly, SAN-based encoder without position embedding fails on both machine translation and our WRD task, indicating the necessity of position encoding on learning word order. It is encourage to see that SAN yields higher BLEU score and detection accuracy than ``DiSAN-Pos\_Emb'' in machine translation scenario. It means that position embedding is more suitable on capture word order information for machine translation than modeling recurrence for SAN. Considering both two scenarios, DiSAN-based encoders achieve comparable detection accuracies to the best models, revealing its effectiveness and universality on learning word order.     



\subsection{Analysis} 
\label{sec:ana}

In response to above results, we provide further analyses to verify our hypothesis on NMT encoders. We discuss three questions in this section: 1) Does learning objective indeed affect the extracting of word order information; 2) How SAN derives word order information from position encoding; and 3) Whether more word order information retained is useful for machine translation. 

\paragraph{Accuracy According to Distance}
We further investigate the accuracy of WRD task according to various distance between the positions of word is popped out and inserted. 
As shown in Figure~\ref{fig:detect-dis} (a), WRD encoders marginally reduce the performance with the increasing of distances. 
However, this kind of stability is destroyed when we pre-train each encoder with a learning objective of machine translation. 
As seen in Figure~\ref{fig:detect-dis} (b) and (c), the performance of pre-trained NMT encoders obviously became worse on long-distance cases across language pairs and model variants. 
This is consistent with prior observation on NMT systems that both RNN and SAN fail to fully capture long-distance dependencies~\cite{tai-socher-manning:2015:ACL-IJCNLP,yang2017towards,Tang:2018:EMNLP}. 

Regarding to information bottleneck principle~\cite{tishby2015deep,alemi2016deep}, our NMT models are trained to maximally maintain the relevant information between source and target, while abandon irrelevant features in the source sentence, e.g. portion of word order information.  Different NLP tasks have distinct requirements on linguistic information~\cite{conneau2018you}. 
For machine translation, the local patterns (e.g. phrases) matter more~\cite{Luong:2015:EMNLP,Yang:2018:EMNLP,Yang:2019:NAACL}, while long-distance word order information plays a relatively trivial role in understanding the meaning of a source sentence. 
 Recent studies also pointed out that abandoning irrelevant features in source sentence benefits to some downstream NLP tasks~\cite{leirationalizing,yu2017learning,shen2018reinforced}. 
An immediate consequence of such kind of data process inequality~\cite{schumacher1996quantum} is that information about word order that is lost in encoder cannot be recovered in the detector, and consequently drops the performance on our WRD task.
The results verified that the learning objective indeed affects more on learning word order information than model architecture in our case. 


\paragraph{Accuracy According to Layer}
Several researchers may doubt that the parallel structure of SAN may lead to failure on capturing word order information at higher layers, since the position embeddings are merely injected at the input layer. 
Accordingly, we further probe the representations at each layer on our WRD task to explore how does SAN learn word order information. 
As seen in Figure~\ref{fig:MT-layer}, SAN achieves better performance than other NMT encoders on the proposed WRD tasks across almost all the layers.
The result dispels the doubt on the inability of position encoding and confirms the speculation by~\newcite{Vaswani:2017:NIPS} and \newcite{Shaw:2018:NAACL} who suggested that SAN can  profit from the use of residual network which propagates the positional information to higher layers. 
Moreover, both SAN and RNN gradually increase their performance on our task with the stacking of layers. The same tendency demonstrates that position encoding is able to provide same learning manner to that of recurrent structure with respect to word order for SAN. Both the results confirm the strength of position encoding to bring word order properties into SAN.

We strove to come up with the reason why SAN captured even more word order information in machine translation task. 
~\newcite{yin2017comparative} and ~\newcite{tran2018importance} found that the approach with a recurrence structure (e.g. RNN) has an easier time learning syntactic information than that of models with a parallel structure (e.g. CNN, SAN). 
Inspired by their findings, we argue that SAN tries to partially countervail its disadvantage in parallel structure by reserving more word order information, thus to help for the encoding of deeper linguistic properties required by machine translation. 
Recent studies on multi-layer learning shown that 
different layers tend to learn distinct linguistic information~\cite{Peters:2018:NAACL,Raganato:2018:EMNLPWorkshop,Li:2019:NAACL}. 
The better accuracy achieved by SAN across layers indicates that SAN indeed tries to preserve more word order information during the learning of other linguistic properties for translation purpose.





\begin{figure}[t]
\begin{center}
\includegraphics[width=0.36\textwidth]{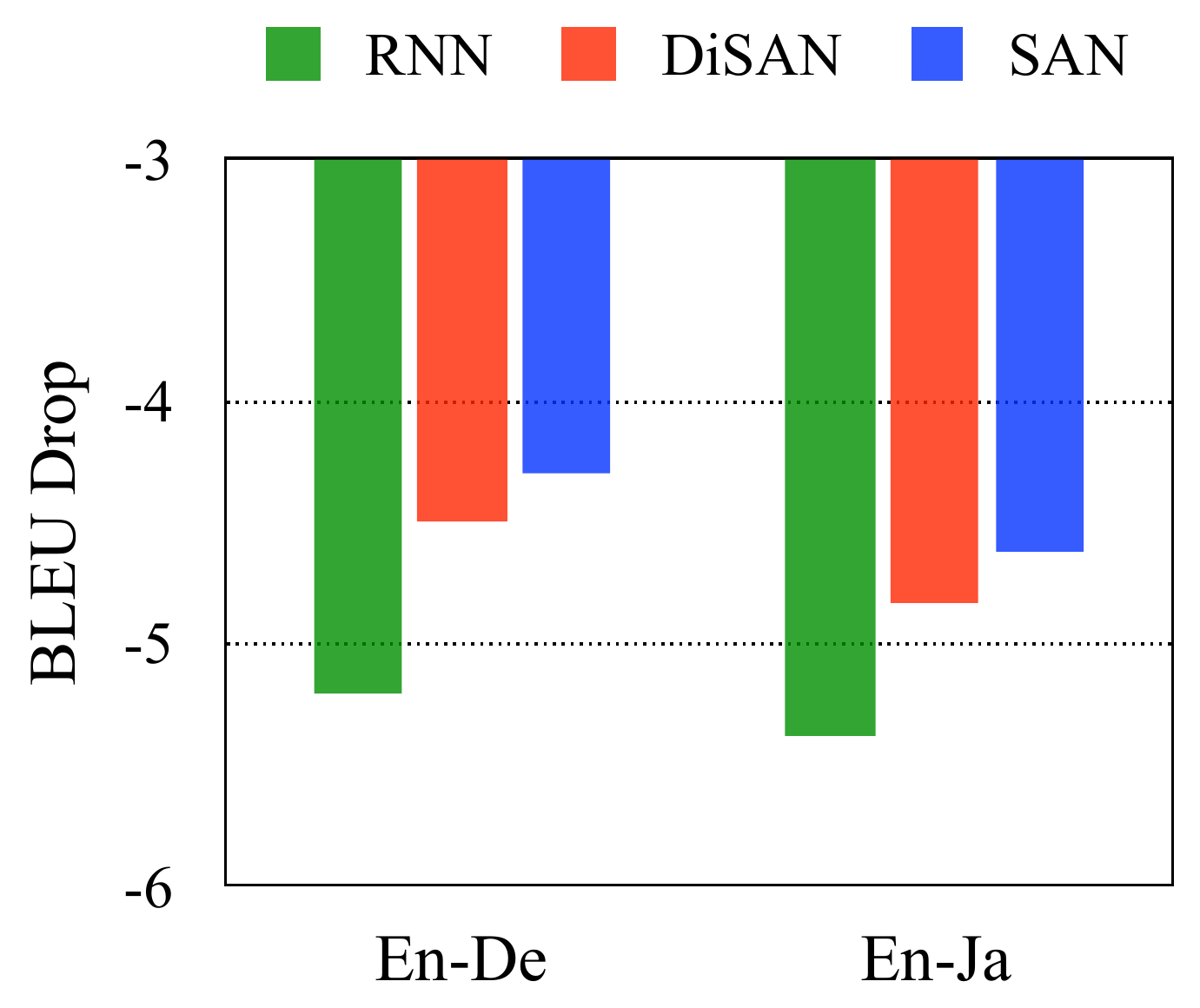}
\caption{The differences of translation performance  when the pre-trained NMT models are fed with the original (``Golden'') and reordered (``Reorder'') source sentences. As seen, SAN and DiSAN perform better on handling noises in terms of erroneous word order.}
\label{fig:gap}
\end{center}
\end{figure}

\paragraph{Effect of Wrong Word Order Noises}
For humans, a small number of erroneous word orders in a sentence usually does not affect the comprehension. For example, we can understand the meaning of English sentence ``Dropped the boy the ball.'', despite its erroneous word order. It is intriguing whether NMT model has the ability to tackle the wrong order noises. As a results, we make erroneous word order noises on English-German development set by moving one word to another position, and evaluate the drop of the translation quality of each model. As listed in Figure~\ref{fig:gap}, SAN and DiSAN yield less drops on translation quality than their RNN counterpart, demonstrating the effectiveness of self-attention on ablating wrong order noises.  
We attribute this to the fact that models (e.g. RNN-based models) will not learn to be robust to errors since they are never observed~\cite{sperber2017toward,cheng2018towards}. On the contrary, since SAN-based NMT encoder is good at recognizing and reserving anomalous word order information under NMT context, it may raise the ability of decoder on handling noises occurred in the training set, thus to be more robust in translating sentences with anomalous word order.

\section{Related Work}

\paragraph{Exploring Properties of SAN}
 {SAN has yielded strong empirical performance in a variety of NLP tasks~\cite{Vaswani:2017:NIPS,tan2018deep,li2018multi,devlin2018bert}.
 In response to these impressive results, several studies have emerged with the goal of understanding SAN on many properties.} 
For example, \newcite{tran2018importance} compared SAN and RNN on language inference tasks, and pointed out that SAN is weak at learning  hierarchical structure than its RNN counterpart. 
Moreover,~\newcite{Tang:2018:EMNLP} conducted experiments on subject-verb agreement and word sense disambiguation tasks. They found that SAN is good at extracting semantic properties, while underperforms RNN on capturing long-distance dependencies. This is in contrast to our intuition that SAN is good at capturing long-distance dependencies. 
 In this work, we focus on exploring the ability of SAN on modeling word order information. 
 
\paragraph{Probing Task on Word Order}
 {To open the black box of networks, probing task is used as a first step which facilitates comparing different models on a much finer-grained level. Most work has focused on probing fixed-sentence encoders, e.g. sentence embedding~\cite{adi2016fine,conneau2018you}. Among them,} 
\newcite{adi2016fine} and \newcite{conneau2018you} introduced to probe the sensitivity to legal word orders by detecting whether there exists a pair of permuted word in a sentence by giving its sentence embedding. However, analysis on  sentence encodings may introduce confounds, making it difficult to infer whether the relevant information is encoded within the specific position of interest or rather inferred from diffuse information elsewhere in the sentence~\cite{tenney2018you}. In this study,  we directly probe the token representations
for word- and phrase-level properties, which has been widely used for probing token-level representations learned in neural machine translation systems, e.g. part-of-speech, semantic tags, morphology as well as constituent structure \cite{shi-padhi-knight:2016:EMNLP2016,belinkov2017neural,blevins2018deep}. 

\section{Conclusion}
In this paper, we introduce a novel {\em word reordering detection} task which can probe the ability of a model to extract word order information. With the help of the proposed task, we evaluate RNN, SAN and DiSAN upon Transformer framework to empirically test the theoretical claims that SAN lacks the ability to learn word order. The results reveal that RNN and DiSAN exactly perform better than SAN on extracting word order information in the case they are trained individually for our task. However, there is no evidence that SAN learns less word order information under the machine translation context. 

Our further analyses for the encoders pre-trained on the NMT data suggest that 1) the learning objective sometimes plays a crucial role on learning a specific feature (e.g. word order) in a downstream NLP task; 
2) modeling recurrence is universally-effective to learn word order information for SAN; 
and 3) RNN is more sensitive on erroneous word order noises in machine translation system. 
These observations facilitate the understanding of different tasks and model architectures in finer-grained level, rather than merely in overall score (e.g. BLEU). 
{As our approach is not limited to the NMT encoders, it is also interesting to explore how do the models trained on other NLP tasks learn word order information.}

 
 
\section*{Acknowledgments}

The work was partly supported by the National Natural Science Foundation of China (Grant No. 61672555), the Joint Project of Macao Science and Technology Development Fund and National Natural Science Foundation of China (Grant No. 045/2017/AFJ) and the Multi-Year Research Grant from the University of Macau (Grant No. MYRG2017-00087-FST).
We thank the anonymous reviewers for their insightful comments.

\balance
\bibliography{acl2019}
\bibliographystyle{acl_natbib}

\end{document}